# Complex System Diagnostics Using a Knowledge Graph-Informed and Large Language Model-Enhanced Framework

Saman Marandi[1], Yu-Shu Hu[2], Mohammad Modarres[1]

[1]Center for Risk and Reliability University of Maryland, MD, USA; [2]DML Inc., Hsinchu, Taiwan

**Corresponding Author:** `smarandi@umd.edu`

## Abstract

In this paper, we present a novel diagnostic framework that integrates Knowledge Graphs (KGs) and Large Language Models (LLMs) to support system diagnostics in high-reliability systems such as nuclear power plants. Traditional diagnostic modeling struggles when systems become too complex, making functional modeling a more attractive approach. Our approach introduces a diagnostic framework grounded in the functional modeling principles of the Dynamic Master Logic (DML) model. It incorporates two coordinated LLM components, including an LLM-based workflow for automated construction of DML logic from system documentation and an LLM agent that facilitates interactive diagnostics. The generated logic is encoded into a structured KG, referred to as KG-DML, which supports hierarchical fault reasoning. Expert knowledge or operational data can also be incorporated to refine the model's precision and diagnostic depth. In the interaction phase, users submit natural language queries, which are interpreted by the LLM agent. The agent selects appropriate tools for structured reasoning, including upward and downward propagation across the KG-DML. Rather than embedding KG content into every prompt, the LLM agent distinguishes between diagnostic and interpretive tasks. For diagnostics, the agent selects and executes external tools that perform structured KG reasoning. For general queries, a Graph-based Retrieval-Augmented Generation (Graph-RAG) approach is used, retrieving relevant KG segments and embedding them into the prompt to generate natural explanations. A case study on an auxiliary feedwater system demonstrated the framework's effectiveness, with over 90% accuracy in key elements and consistent tool and argument extraction, supporting its use in safety-critical diagnostics.

## 1. Introduction

The analysis of complex engineered systems, particularly those in high-reliability and safety-critical industries such as nuclear power plants, requires systematic approaches to assess system integrity, reliability, and performance. Traditional diagnostic tools have predominantly relied on event-based modeling, where the outcomes of specific failure pathways, faults or abnormal initiating events are analyzed. Although effective in some domains, event-based approaches become extremely complex, incomplete and inadequate when applied to complex, interconnected systems with numerous interdependencies. As the scale and complexity of systems increase, functional modeling approaches are considered more suitable, where the emphasis is placed on understanding the roles, dependencies, and contributions of system components toward achieving primary system objectives. Functional modeling involves the development of system models based on the actions and relationships of their constituent parts [1], [2]. These models are inherently hierarchical, enabling a systematic decomposition of the system into goals, functions, and subfunctions. One important framework within this category is the Dynamic Master Logic (DML) model, introduced by Hu and Modarres [3], which represents system behavior through a structured hierarchy linking functional objectives to underlying structures. By organizing systems into functional and structural layers, DML enables critical causal pathways, system interdependencies, and fault propagation mechanisms to be systematically analyzed. This hierarchical framework supports the tracing of failures from system-level objectives down to elemental components, offering a powerful tool for diagnostic analysis. Although DML models provide a robust framework for system diagnostics, their construction, maintenance, and interpretation require significant manual effort and extensive domain expertise. As system complexity grows, the burden associated with constructing and interacting with large DML models increases accordingly. In response to these challenges, powerful Artificial Intelligence (AI) tools such as Large Language Models (LLMs) [4] have been recognized as offering opportunities to enhance the development, interaction, and usability of functional models for diagnosis of faults and failures in complex engineering systems. LLMs have demonstrated strong capabilities in natural language understanding, summarization, and reasoning across a wide range of domains. However, certain limitations such as hallucination and restricted domain-specific reasoning have been observed. To address these challenges and improve diagnostic reliability, structured knowledge representations such as Knowledge Graphs (KGs) [5] can be used alongside LLMs to enable more consistent and interpretable reasoning by guiding diagnostic logic through external tools rather than unconstrained language generation.

The integration of LLMs with KGs provides a mechanism to enhance the accessibility, organization, and interpretability of complex functional models. By grounding LLM interactions in verified domain knowledge, the risks associated with hallucination can be reduced and the transparency of diagnostic outputs can be improved. Given the increasing complexity of engineered systems and the need for scalable, interpretable diagnostic tools, there is a strong motivation to explore AI-driven frameworks that combine domain knowledge with language-based reasoning. In this paper, an approach is proposed that leverages LLMs and KGs to support and enhance interaction with DML models, to reduce manual effort, improve transparency, and facilitate more efficient fault analysis in safety-critical applications.

The remainder of this paper is organized as follows. Section 2 reviews related work on DML modeling, large language models, and the integration of KGs in diagnostic systems. Section 3 provides an overview of the research approach. Section 4 presents the proposed diagnostic framework, detailing both the model construction and interaction phases. Section 5 describes a case study involving an auxiliary feedwater system in a nuclear power plant, illustrating the application of the framework. Section 6 reports the evaluation results from both the model construction and interaction components, based on repeated runs using the case study. Finally, Section 7 concludes the paper with key findings and discusses directions for future work.

## 2. Background

### 2.1. From Traditional Diagnostics to Functional Modeling

Traditional diagnostic approaches often rely on modeling discrete events, failure modes, or symptom triggers. These include methods like Fault Tree Analysis (FTA) [6], [7], Event Trees (ET) [8], [9], and rule-based [10], [11] expert systems that map observed events to likely faults. Such models focus on specific failure events and their consequences. While these event-driven models can effectively represent known failure sequences, they typically require an exhaustive list of fault-event combinations, invariably making them incomplete. If a particular sequence is not anticipated during model development, the system may fail to diagnose it. Moreover, as systems grow in complexity, tracing failure pathways become increasingly infeasible using event-based logic alone. Functional modeling reflects a designer's intent by capturing system goals and functions rather than enumerating events. These models encode the functional expectations and logical dependencies of the system, enabling the detection of faults based on deviations from expected behavior. These models are hierarchical, representing goals, functions, subfunctions, and supporting structures. They emphasize how components contribute to overall functions, enabling reasoning about failures in terms of lost or abnormal functionality, such as a pump failing to provide flow or a sensor failing to deliver information, rather than focusing solely on specific failure events [12]. Functional approaches address several limitations of event-based models. First, they support completeness by enabling a structured analysis of the system's functional architecture. This allows for the identification of faults that may cause the loss of required functions, even during the design or concept stage before any failure data is available. Second, functional models naturally handle complex, multi-fault scenarios. Since they capture interactions through functional dependencies, they can reason about multiple simultaneous failures or cascading effects without relying on every combination explicitly. Third, they promote generality. The same functional model can be applied throughout the system's life cycle and across various analysis tasks. These include design-level Failure Mode and Effects Analysis (FMEA), runtime diagnosis, and what-if analysis. This flexibility is possible because the model abstracts away from specific event sequences and instead focuses on invariant functional relationships [13].

*2.2. DML Model Applications*

For system-level modeling, DML models are success logics in the form of powerful hierarchies to represent system knowledge [3], [14], [15], [16]. This hierarchical representation becomes particularly valuable in safety-critical domains. System safety focuses on applying engineering principles, standards, and techniques to minimize risk while ensuring a system remains effective, reliable, and cost-efficient throughout its life cycle. Using a DML, the degree of success (or failure), approximate full-scale system physical values, and transition effects amongst components can be analyzed. DML provides an effective model for describing the causal effects of failures or disturbances in complex systems [17]. Figure 1 conceptually illustrates the DML framework, highlighting the hierarchical decomposition from objectives to basic elements and the interdependencies between functional and structural. The functional hierarchy (top-down) moves from objectives to functions and sub-functions, capturing the system's purpose and behavior. The structural hierarchy decomposes the system into elements and basic components, reflecting its physical or logical composition. Arrows indicate causal ("Why–How") and compositional ("Part-of") relationships, highlighting interdependencies. This structure supports systematic analysis of complex systems by linking high-level goals to low-level elements.

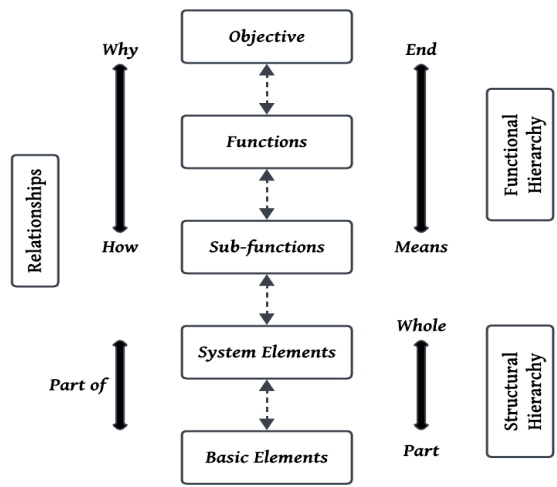

Figure 1. Conceptual DML Model [15]

Two key causal relationships can be extracted from DML. The first involves determining the ultimate effect of a failure, and the second involves identifying the paths through which a function can be achieved or a subsystem can successfully operate. It applies reductionist principles, where qualities represent functions and goals, while objects and relationships are structured through success trees and logical modeling including Boolean, physical, and fuzzy logic [14]. This integration enhances DML's ability to model time-dependent behaviors and fault propagation within complex systems. This model has been applied to various applications of modeling. In nuclear power plants, DML has been used to model Direct Containment Heating in Pressurized Water Reactor [16]. In renewable energy, it has supported reliability analysis of geared wind turbines [18]. DML has also proven effective in analyzing interactions between hardware, software, and human elements in cyber-physical systems [19], [20]. In the aerospace sector, it has been used to identify critical points in system reliability [21]. Additionally, DML has supported quality assurance across the software development life cycle [22].

It is important to note that several naming conventions have been used to describe what is fundamentally a single family of DML models with common underlying principles. The earliest form of this modeling approach was the Goal Tree Success Tree (GTST) [12], which laid the foundation for functional reasoning in complex systems. This was followed by the development of the Master Plant Logic Diagram (MPLD) [23], [24], originally created for use in nuclear power plants to represent plant logic and support reliability assessment. Over time, MPLD evolved into several extensions and refinements. A prominent variant is the GTST with Master Logic Diagram (MLD), referred to as GTST-MLD [17], [22]. This version combines a functional hierarchy, represented by the GTST with a structural model, represented by the MLD that captures component relationships and dependencies, supporting both dynamic and static system representations. Another well-established form is the Dynamic Master Logic Diagram (DMLD) [14] which emphasizes the modeling of uncertain, evolving, and time-dependent behaviors in complex systems. While terminology may differ depending on modeling emphasis, these approaches are functionally equivalent and share the common goal of representing system logic for diagnostic reasoning and reliability analysis. Throughout this paper, this family of models is collectively referred to as DML.

### 2.3. LLMs and KGs in Fault Diagnostics

Recent advancements in LLMs have accelerated the use of fault diagnostics, particularly in industrial and safety-critical domains. Traditional fault diagnosis methods rely heavily on rule-based models or historical fault logs, which require extensive domain expertise and are often inefficient in handling complex, multi-layered system interactions. Integrating LLMs with diagnostic methods makes timely detection of faults more practical and easier to scale by enabling an understanding of fault causes and providing decision support through NLP. As demonstrated in [25], a diagnostic model was used to detect faults in a nuclear power plant, while an LLM acted as an interface to explain fault conditions and their potential causes to human operators, enhancing situational awareness and response.

A study introduced the method of FD-LLM [26], framed machine fault diagnosis as a text classification task by converting vibration signals into textual token sequences or summaries. Fine-tuned LLaMA models were employed, achieving superior performance compared to traditional deep learning models and demonstrating that LLMs can effectively process non-textual diagnostic information when appropriately structured. Improvements were reported in accurately identifying fault root causes from symptom descriptions, highlighting the potential of LLMs for natural language-based symptom interpretation and fault retrieval. To enhance the reliability and reasoning capabilities of LLM-based diagnostics, several studies have proposed integration with KGs. One approach, Root-KGD [27], combined industrial process data with a structured KG representing system topology and causal dependencies, enabling more accurate root cause failure identification by guiding LLM reasoning through formalized domain knowledge. In the context of CNC machine fault diagnosis [28], a KG-embedded LLM architecture was proposed. A machining-process KG was automatically constructed from maintenance records, and its structured information was embedded into the LLM to support fault classification and provide explainable fault identification through natural language outputs.

Integration of KGs into fault diagnostics has been explored to address the need for structured causal reasoning. A multi-level KG was developed to represent rotating machinery faults [29], and Bayesian inference was applied to trace symptom-cause pathways within the graph, achieving a 91.1% diagnostic accuracy under missing data conditions. This demonstrated that the structured representation of symptom-cause relationships within a KG can enhance diagnostic robustness. The combination of structured knowledge retrieval and LLM reasoning enabled interpretable and accurate fault diagnosis based on free-form symptom descriptions. The combination of LLMs with KGs has also demonstrated enhanced fault reasoning capabilities.

A hybrid model for aviation assembly diagnostics achieved 98.5% accuracy in fault localization and troubleshooting through subgraph-based reasoning [30]. In vehicle fault diagnostics, a KG-driven analysis system was developed that maps error codes and system alerts to potential failure causes [31]. This system leveraged LLMs to process unstructured diagnostic data, such as error logs and maintenance reports, transforming it into structured representations within a KG. A reasoning framework was introduced to infer root causes by linking symptoms to failure mechanisms stored in the KG. In [32], a KG-based in-context learning approach was proposed to enhance fault diagnosis in industrial sensor networks. A domain-specific KG was constructed to encode expert knowledge, and a long-length entity similarity retrieval mechanism was used to select relevant knowledge, which was then supplied to a large language model for causal reasoning over fault symptom text. The method demonstrated improved fault localization accuracy and enhanced the interpretability of diagnostic outputs compared to traditional LLM approaches.

These studies demonstrated that incorporating LLM-driven fault reasoning improved diagnostic accuracy and reduced troubleshooting time compared to traditional rule-based or statistical models. Structuring fault data within a KG provided traceable and explainable reasoning paths, assisting engineers and technicians in understanding failure causes. Collectively, these studies illustrate the growing role of LLMs and KGs in fault diagnostics. LLMs enable flexible interpretation of symptom descriptions and support natural language interaction, while KGs structure domain-specific knowledge to guide reasoning processes. Their integration shows promise for enhancing fault retrieval and root cause analysis across various technical domains.

## 3. Research Overview

As discussed in Section 2, although LLMs and KGs show promise for assisting with system diagnostics from textual documentation, their role in structured functional decomposition is not fully developed. Similarly, KGs are primarily used for retrieving fault information rather than representing system logic or dependencies, and few studies have explored integrating computational reasoning layers to enable real-time diagnostic analysis. To address these gaps, this paper introduces an LLM-informed, KG-based diagnostic framework for constructing and using the DML model derived from system design and operational documents. The framework leverages LLMs to support diagnostics, reliability assessment, and decision-making for a specific engineering system. It has three main objectives:

a. Automate the generation of scalable functional models by extracting structured relationships from system documentation, including system descriptions and specifications.
b. Enable interactive, hierarchical fault analysis through natural language-driven upward and downward reasoning. Upward reasoning evaluates how individual component failures affect higher-level system functions, while downward reasoning explores which functional paths and component conditions must be satisfied to maintain or restore system objectives.
c. Support interpretive analysis of system goals, functions, and dependencies by leveraging Graph-based Retrieval-Augmented Generation (Graph-RAG) to retrieve and contextualize relevant segments of the KG in response to user queries.

## 4. Proposed Approach

Figure 3 presents the LLM-Informed Diagnostic Framework, which consists of two main stages: model construction and model interaction. In the model construction stage, system descriptions are processed by an LLM-based workflow to extract DML logic, which organizes the system into hierarchical functions, components and their relationships. This representation is then used to build a KG, which serves as the core system model. The KG can be further enhanced by incorporating expert knowledge and real-time operational data processed through Machine Learning (ML) or Deep Learning (DL) models, allowing it to infer and reflect system states and conditions dynamically. In the diagnostic model interaction section, an LLM agent would determine the intention of the user and would invoke one of the tools available to it based on the query to execute. These tools are used to trace the KG to generate diagnostic insights. The results would be communicated to the user by the LLM.

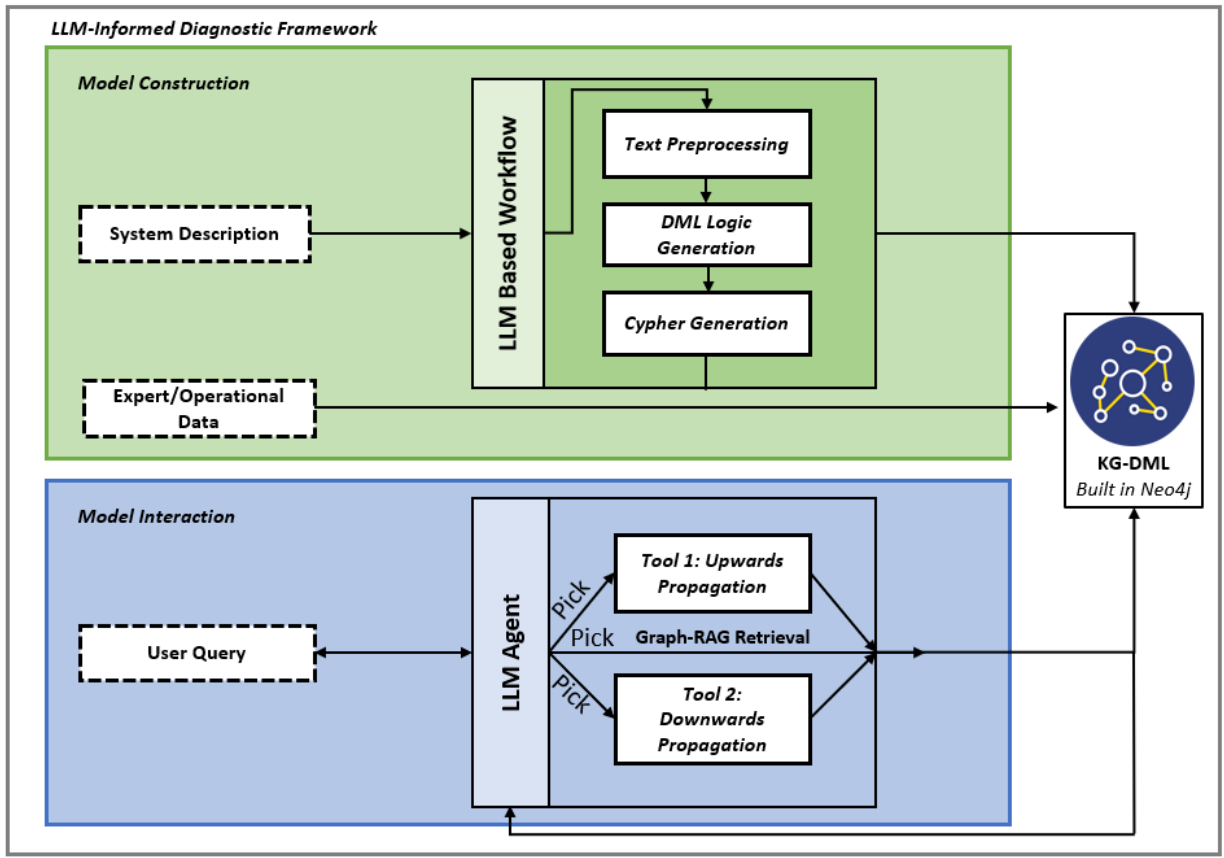

Figure 2. LLM-Informed Diagnostic Framework

*4.1. Model Construction*

The development of the diagnostic model begins with processing textual system information, including manuals, system descriptions, and technical documentation. Preprocessing techniques such as text summarization and LLM-based Named Entity Recognition (NER) are applied to extract key details about system components, functions, and dependencies. These extracted elements are then used to derive a DML hierarchical structure that organizes the system into goals, functions, subfunctions, and components. This structured logic is subsequently translated into Cypher code, the query language used to construct the KG that represents the system's components and functional relationships. The resulting KG provides a structured repository to support querying and reasoning for diagnostics and decision-making. In this research, the KG is deployed using Neo4j [33], a graph database platform that represents entities and their interconnections as property graphs, with attributes stored as node and relationship properties. Cypher enables the definition of system dependencies within the KG, linking components to their success conditions and subfunctions to their parent functions. All language-based tasks in this framework were performed using OpenAI's GPT-4 model through the ChatGPT API.

*4.2. Model Interaction*

Once the model is built, it needs to be used to generate diagnostic insights. This framework aims to go beyond simple queries by enabling deeper system analysis, addressing the fundamental diagnostic questions:

- *What* is happening? The model identifies current system conditions. For example, it can determine which components are degrading and which system functions are at risk.
- *Why* is it happening? By tracing system dependencies, the model identifies apparent root causes of failures and analyzes contributing factors.
- *How* will it impact the system? The model assesses failure propagation and risk severity, predicting how component failure will affect system operations.

To address these questions, the model enables cause-and-effect reasoning, allowing users to explore system behavior beyond basic retrieval. Instead of just fetching stored information, it supports queries such as:

- If certain components fail, how will it affect the overall system?
- What conditions must be met for a specific function to succeed?
- Which components are essential to maintaining system functionality?

To support this process, a set of predefined functions is made available to the LLM agent as tools for interfacing with the KG. These tools enable structured upward and downward tracing to analyze system dependencies and generate diagnostic insights. When a user submits a query, the LLM interprets the intent and selects the appropriate tool to execute. The output is then passed back to the LLM, which produces a human-readable explanation. In addition to tool-based diagnostics, the model supports general system queries, such as explaining the system's hierarchy or functional structure. For these interpretive queries, the framework employs a Graph-RAG approach where relevant graph segments, such as goals or functions, are retrieved and embedded into the LLM's prompt to support natural language generation. Both approaches rely on the KG, but in complementary ways. Diagnostic tools perform structured graph traversals to compute logic-based results, while Graph-RAG enables the LLM to produce contextual explanations based on retrieved subgraphs. This dual strategy ensures the KG remains central to both reasoning and explanation.

## 5. Case Study

The proposed system illustrated in the Piping and Instrumentation Diagram (P&ID) of Figure 3 has been used as a case study to implement the proposed LLM-informed diagnostic framework. This P&ID represents a simplified auxiliary feedwater system of a nuclear power plant. The auxiliary feedwater system in a pressurized light water nuclear plant ensures the safe and efficient supply of emergency cooling water to the steam generators during unexpected plant transients. The system is structured with the main goal defined as "Ensure safe and effective operation of the system". This goal is supported by four primary functions: "Supply Feedwater", "Control Water Flow", "Manage System Integration and Response", and "Provide Emergency and Automated Response".

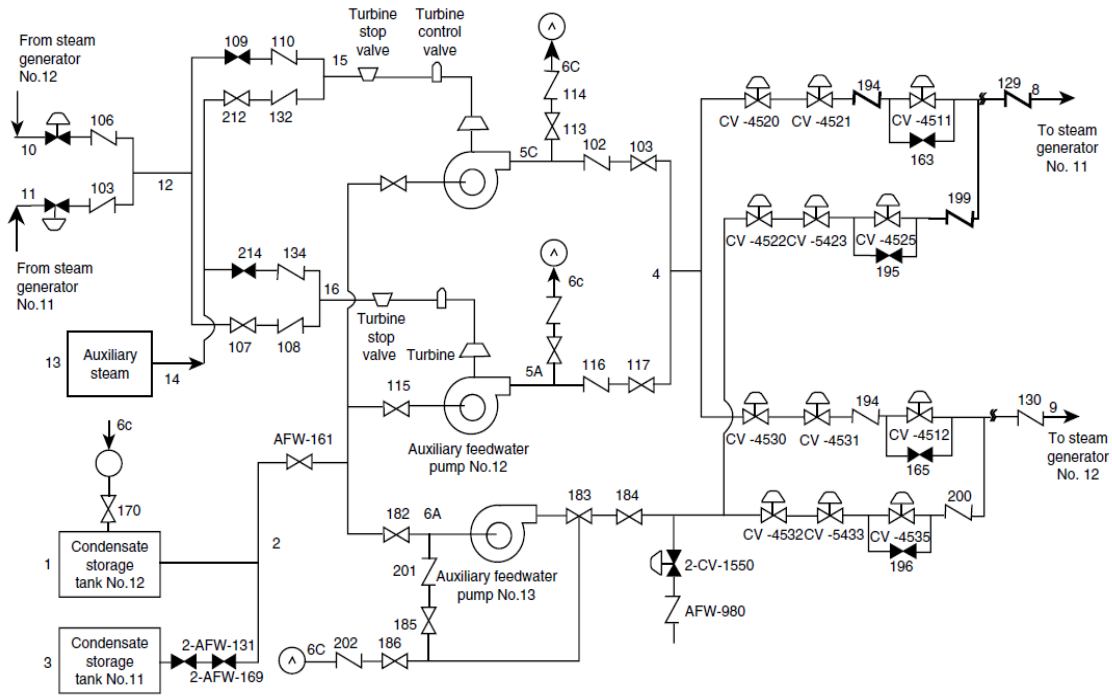

Figure 3. P&ID of Simplified Auxiliary Feedwater System [34]

### 5.1. Model Construction

To construct a KG representing DML logic from a system description (which included only major components and functions), a structured prompt chaining workflow was implemented, with each stage handled by a dedicated LLM call. The hierarchy follows the DML structure, starting with a high-level goal and breaking down into functions, subfunctions, and components, each linked to success conditions. Logical relationships are shown using binary AND or OR logic gates, which define how lower-level elements contribute to achieving higher-level objectives.

The workflow begins with an initial LLM call that summarizes the system description and extracts goals, functions, subfunctions, components, and success conditions. The result is passed to a second LLM that converts this information into a structured JSON format aligned with the DML hierarchy. A third LLM then transforms the JSON into Cypher queries for KG construction. Each LLM call is followed by a gate, implemented as another LLM, that validates the output before the next stage proceeds. If validation fails, the workflow routes the input back to the relevant LLM for revision. The first gate checks for missing or incomplete information in the summary, including vague goals, incomplete function chains, or missing success criteria. The second gate validates the JSON structure by checking key formatting, nesting, and logical gate consistency. The third gate examines the generated Cypher queries to ensure they are syntactically correct.

This gated prompt chaining design improves consistency, filters out errors early, and manages variability in LLM outputs. It is especially effective when task steps are clearly defined and expected outputs are explicitly structured. Figure 4 illustrates the prompt chaining workflow described above, showing the sequential LLM tasks and validation gates leading from the system description to the final KG-DML output. Each task is followed by an LLM-based gate, which ensures the correctness of the output before advancing to the next stage. Feedback loops are included to allow correction and regeneration when validation fails. The specific prompts used for each LLM call in this workflow are provided in the Appendix to this paper.

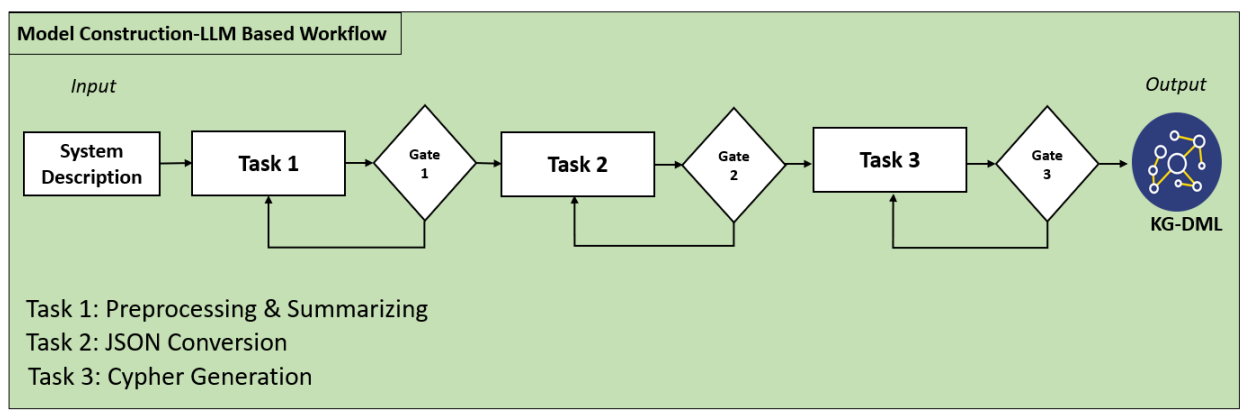

Figure 4. Model Construction Implementation Through LLM-Based Workflow

The KG representing the DML logic is structured hierarchically, starting from a high-level goal and descending through functions, subfunctions, components, and finally success conditions. Logical gates, such as AND or OR, define how each level contributes to achieving the level above. A goal may be achieved by multiple functions, each of which depends on one or more subfunctions that require specific components to operate successfully. At the lowest level of the hierarchy, components are connected to success conditions through additional gates. Success conditions reflect observable or measurable outcomes that confirm whether a component is performing as intended. Attributes are stored within the nodes themselves and may include expert knowledge or information derived from ML or DL models based on operational data or manual inspections.

For example, a component such as a turbine-driven pump may contain attributes indicating the probability of being in various states, such as operational, degraded, or failed. Attributes may also be present in higher-level nodes, such as functions or goals. Their role and interpretation will be discussed in the Model Interaction (next section). This hierarchical structure supports reasoning, traceability, and consistency throughout the KG-DML representation. Figure 5 illustrates how the DML model is represented within the KG. For example, the subfunction "Manage Condensation Tanks" is fulfilled only if all three Condensation Storage Tanks (CSTs) operate successfully, as defined by an AND gate. Each tank must meet two success conditions: maintaining an appropriate water level and ensuring the absence of excessive sediment. This same hierarchical logic applies when tracing the model upward through functions and system-level goals.

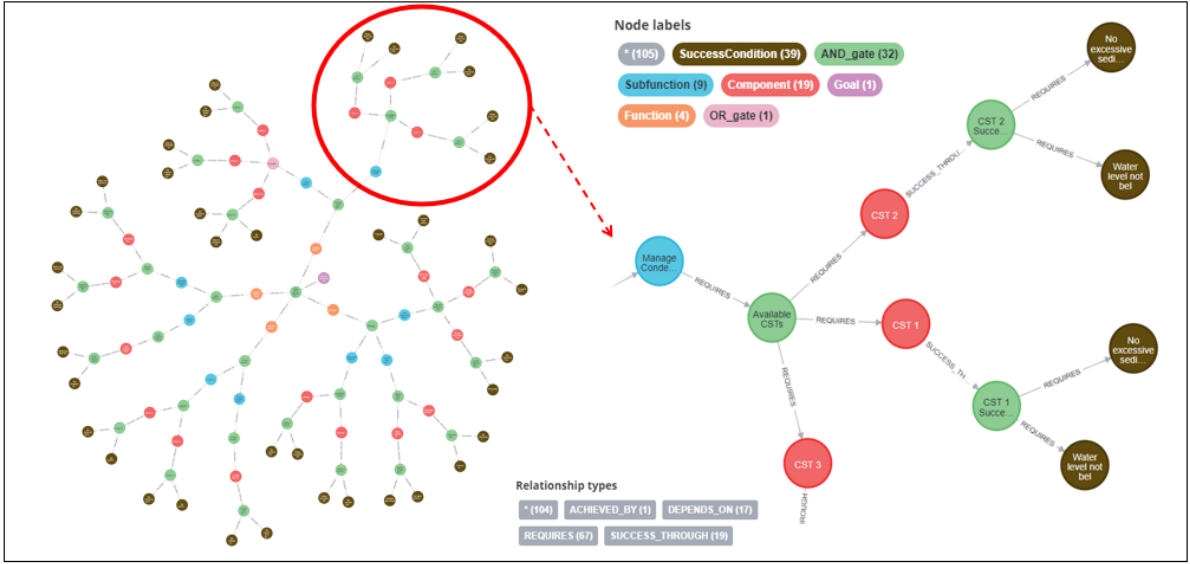

Figure 5. KG Reflecting the DML Model

## 5.2. Model Interaction

As shown in Figure 2, model interaction begins when a user submits a natural language query to the system. The LLM agent interprets the query and selects from a set of predefined diagnostic functions, available to it as tools. These tools perform specialized tasks such as upward fault tracing and generation of success path sets. Each tool is implemented as an external code module that analyzes the KG based on the logical structure derived from the model. The LLM uses these tools to carry out structured reasoning over the graph and generate diagnostic insights. To enable accurate selection, the agent was fine-tuned on a dataset consisting of diverse user queries paired with their corresponding tool calls. This included both diagnostic queries requiring tool invocation and interpretive queries requiring Cypher query generation for KG retrieval. This dataset was manually constructed to include multiple phrasings, semantic variations, and tones in which users might pose the same diagnostic intent. These examples were then used to guide the fine-tuning process so that the model learns to map a wide range of natural language inputs to the appropriate tool.

In the implemented tool for upward propagation, the success probability for each success condition $j$ associated with a component is first evaluated. This is done using Equation 1, which computes the probability as a weighted sum over the component's possible operational states. Each term combines the likelihood of the component being in state $i$ with the probability that it fulfills success condition $j$ in that state. A state refers to a possible condition of a component, such as operational, degraded, or failed. Each state influences the component's ability to fulfill its associated success conditions. For example, consider a CST in the auxiliary feedwater system. One possible state of the CST is "failed", which may be inferred from sensor data. The success condition for the CST could be defined as "maintains sufficient water level for feedwater supply." If operational data indicates a high probability that the CST is in a failed state, the likelihood of satisfying this success condition would be correspondingly low. This affects the overall success probability of the subfunction "Manage Condensation Tanks," which depends on all CSTs through an AND gate. Thus, the failure of even one CST can reduce the success probability of the higher-level function and system goal.

$$P(Success_j|Data) = \sum_{i=1}^{N} P(Success_j|State_i)P(State_i|Data) \qquad (1)$$

- $N$: Total number of operational states for a component.
- $P(Success_j|State_i)$: The probability of success for the success condition $j$. This reflects how likely the component will fulfill the success condition $j$ under state $i$.
- $P(State_i|Data)$: The probability of the component being in state $i$ given the data which can be evidence of events or numerical information.

After evaluating $P(Success_j|Data)$ for all success conditions associated with each component in the system, the results are aggregated using logical gates to compute a single success probability for each component. Success probability represents how well the component is fulfilling its intended function. It is based on the combined satisfaction of all defined success conditions. Each condition reflects a specific performance indicator. The aggregation of these conditions provides a quantitative measure of the component's overall operational effectiveness. The KG stores each element of Equation 1 as attributes within component nodes, including both the conditional success probabilities and the state likelihoods derived from data. In the context of engineering diagnostics, this data may include sensor readings (e.g., temperature, pressure, vibration), event logs, failure reports, and maintenance histories. These attributes serve as the basis for upward propagation and are retained in the KG to support traceability and diagnostic reporting. Once a single success probability is determined for each component, these values are propagated upward through the DML hierarchy using additional logical gates. The gates define how component-level probabilities combine to determine the success of associated subfunctions, functions, and ultimately system-level goals. If the success probability of an upper-level node falls below a predefined threshold, the tool considers that node to be impacted. The corresponding logic that performs the probabilistic propagation is captured in the pseudocode shown in Figure 6.

To estimate $P(State_i|Data)$, various strategies can be applied depending on data availability and system characteristics. In the absence of real-time sensor data, these probabilities can be derived from expert judgment, or reliability reports, which provide baseline estimates of failure or degradation likelihoods. These priors can be refined as new operational data becomes available. When numerical indicators such as temperature, pressure, or vibration readings are accessible, ML or DL models trained on historical labeled data can be used to estimate state probabilities more dynamically. In systems requiring continuous monitoring and probabilistic inference under uncertainty, particle filtering techniques may be used. Particle filters apply a sequential Monte Carlo approach to approximate probability distributions using a set of weighted samples, enabling real-time Bayesian inference even in nonlinear or non-Gaussian conditions [35], [36].

For downward propagation, given an upper-level node, the tool traces the KG downward to determine the required paths for achieving that node's success. Using the defined gates, it identifies the necessary dependencies at each level. The path-set generation method determines the minimal components required for system functionality by recursively traversing the KG. Starting from a specified node, the process follows dependencies downward until reaching the Component and Success Condition levels. At each step, the method evaluates the logical dependencies based on the gate type. If an AND gate is present, all dependencies must be met simultaneously, requiring a Cartesian product of the success path-sets from the child nodes to generate valid paths. In contrast, for an OR gate, only one dependency needs to succeed, so the success path-sets from the child nodes are aggregated without combination, representing alternative paths to success. This structured approach ensures that the generated success path-sets accurately reflect the minimal elements necessary to maintain system operability. The approach of downward propagation is formalized by the pseudocode in Figure 7.

```
 1: procedure PROPAGATESUCCESSPROBABILITIES
 2:     for each Component C do
 3:         Retrieve possible states State_1, State_2, ..., State_N
 4:         for each Success Condition j of component C do
 5:             Compute:
```
$$P(\text{Success}_j \mid \text{Data}) = \sum_{i=1}^{N} P(\text{Success}_j \mid \text{State}_i) P(\text{State}_i \mid \text{Data})$$
```
 6:         end for
 7:         Combine success conditions using gate type (e.g., AND, OR)
 8:         Update P_success(C) in KG
 9:     end for
10:     for each Subfunction SF do
11:         Retrieve linked Components and gate type
12:         if gateType = AND then
13:             Compute:
```
$$P_{\text{success}}(SF) = \prod P_{\text{success}}(C)$$
```
14:         else
15:             Compute:
```
$$P_{\text{success}}(SF) = 1 - \prod (1 - P_{\text{success}}(C))$$
```
16:         end if
17:         Update P_success(SF) in KG
18:     end for
19:     for each Function F do
20:         Retrieve linked Subfunctions and gate type
21:         if gateType = AND then
22:             Compute:
23:                 P_success(F) = ∏ P_success(SF)
24:         else
25:             Compute:
```
$$P_{\text{success}}(F) = 1 - \prod (1 - P_{\text{success}}(SF))$$
```
26:         end if
27:         Update P_success(F) in KG
28:     end for
29:     for each Goal G do
30:         Retrieve linked Functions and gate type
31:         if gateType = AND then
32:             Compute:
```
$$P_{\text{success}}(G) = \prod P_{\text{success}}(F)$$
```
33:         else
34:             Compute:
```
$$P_{\text{success}}(G) = 1 - \prod (1 - P_{\text{success}}(F))$$
```
35:         end if
36:         Update P_success(G) in KG
37:     end for
38:     Identify impacted nodes where P_success < threshold
39:     return impacted nodes and probabilities[1]
40: end procedure
```

Figure 6. Upwards Propagation Pseudocode

```
 1: procedure GENERATESUCCESSPATHSETS(nodeType, nodeName)
 2:     if nodeType = Component then
 3:         Retrieve success conditions and gate type for nodeName
 4:         if no success conditions exist then
 5:             return {{nodeName}}                          // Component itself forms a success path
 6:         end if
 7:         if gateType = AND then
 8:             return {successConditions}                   // All conditions must be met
 9:         else
10:             return {{cond} for each cond in successConditions}   // Any condition suffices
11:         end if
12:     end if
13:     Retrieve dependencies and gate type for nodeName
14:     if no dependencies exist then
15:         return {{nodeName}}                              // Base case for nodes without dependencies
16:     end if
17:     Initialize childPathsets ← empty list
18:     for each dependency depName in dependencies do
19:         childPaths ← GENERATESUCCESSPATHSETS(depType, depName)
20:         Append childPaths to childPathsets
21:     end for
22:     if gateType = AND then
23:         Initialize combinedPaths ← empty list
24:         for each combination in Cartesian product of childPathsets do
25:             Append concatenated combination to combinedPaths
26:         end for
27:         return combinedPaths
28:     else
29:         return Flattened list of all childPathsets       // Any dependency suffices
30:     end if
31: end procedure
```

Figure 7. Downward Propagation Pseudocode

*5.3. Interaction Interface*

The diagnostic interface enables natural language interaction between the user and the system, allowing users to explore system behavior and fault scenarios. As illustrated in Figure 8, users can ask questions such as the impact of a specific component failure or how a given function can succeed. When queried about the impact of failure of a CST, the LLM agent invokes the upward propagation tool to trace the impact across the system hierarchy. Because the CSTs are connected through an AND gate, the success of higher-level nodes depends on the simultaneous functionality of all CSTs. Therefore, the failure of even a single CST significantly reduces the probability of success for the related subfunctions, functions, and overall system goals. Conversely, when asked about the success conditions for a function like "Supply Feedwater", the agent employs downward tracing to identify all minimal success paths. Results are returned in a human readable format, supporting transparent and intuitive diagnostic analysis without requiring technical familiarity with the underlying model.

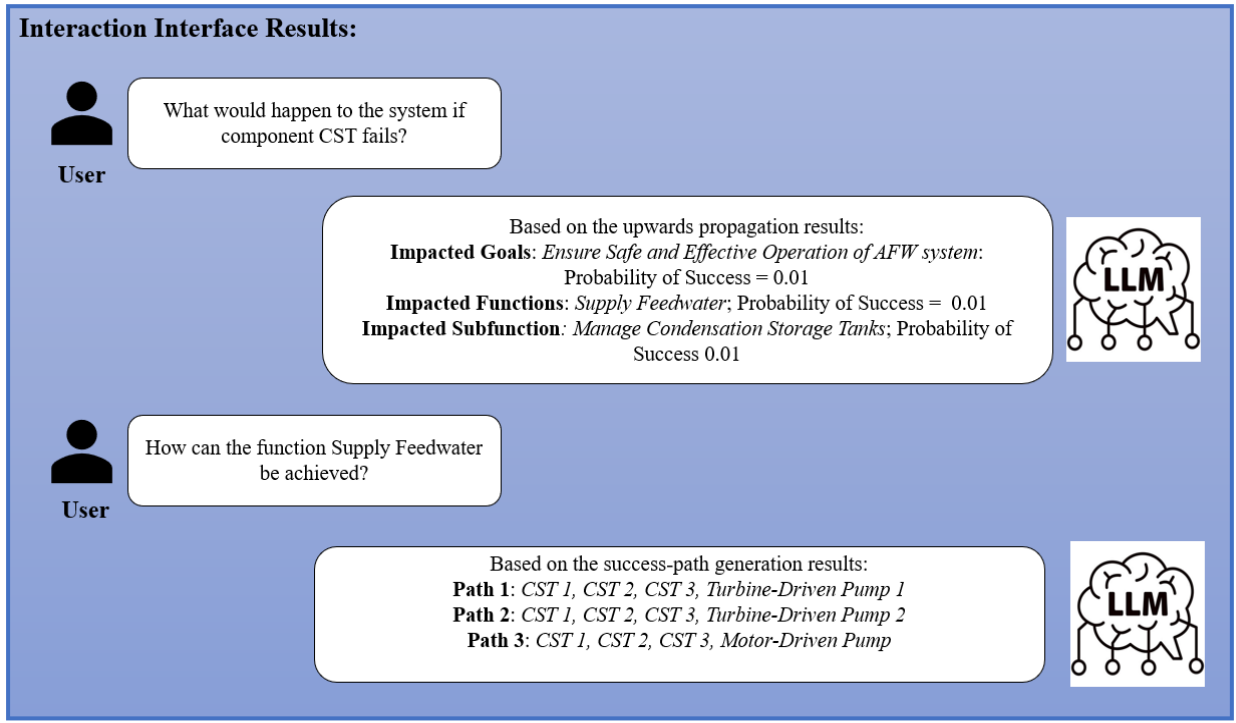

Figure 8. Interaction Interface Example Containing User Sample Questions

## 6. Evaluation

The evaluation of the proposed framework was designed to assess both the structural accuracy of the KG generated from system documentation using the DML hierarchy and the effectiveness of the LLM agent to correctly interpret and respond to diagnostic queries through tool invocation and knowledge retrieval.

### 6.1. KG Validation

To assess the accuracy of the model construction pipeline, we conducted five independent runs using the same system description for the auxiliary feedwater system. In each run, the framework automatically extracted elements of the DML model, including goals, functions, subfunctions, components, success conditions, and logical gates. These outputs were then manually validated. The validation involved examining the resulting KG-DML and cross-referencing its contents with the original system description. An element was considered correctly identified if it was both semantically relevant and structurally consistent with the source material. Elements were labeled as hallucinated if they introduced information that was not present in the documentation or if they misrepresented relationships. The average results across the five runs are summarized in Table 1, which reports the ground truth element counts, the number of correctly extracted elements, the average number of hallucinated elements, and the corresponding extraction accuracy for each model component.

| KG Element | Ground Truth | Avg. Correct | Avg. Hallucinated | Extraction Accuracy (%) |
|---|---|---|---|---|
| Goals | 1 | *1* | *0* | *100.0* |
| Functions | 4 | *3.8* | *0.2* | *95.0* |
| Subfunctions | 9 | *8.6* | *0.4* | *95.6* |
| Components | 19 | *18.2* | *0.8* | *95.8* |
| Logical Gates (AND/OR) | 33 | *30.8* | *2.2* | *93.3* |
| Success Conditions | 39 | *37.2* | *1.8* | *95.4* |

Table 1. Average Validation of KG Elements Across 5 Runs

*6.2. LLM Agent Query Evaluation*

We evaluated the performance of the LLM agent in interpreting natural language queries and selecting the appropriate diagnostic tools or the knowledge-based retrieval mechanism. A test set comprising 60 queries was developed, with queries evenly distributed across three primary task types: upward reasoning, downward reasoning, and explanatory queries. The upward reasoning task involved diagnosing how faults propagate from component-level failures to higher-level system functions. The downward reasoning task focused on identifying the minimal set of components required to achieve a particular function or system goal. Explanatory queries required the agent to retrieve structural or functional information from the KG using a Graph-RAG method, in which relevant nodes and attributes are extracted via automatically generated Cypher queries by the LLM. The evaluation was conducted across five independent runs of the same 60-query dataset to account for variability in LLM outputs. Each query was assessed based on three criteria. These included whether the agent correctly classified the task type, whether the tool or retrieval method was selected appropriately, and whether the extracted arguments or generated Cypher queries were correct. Argument extraction and Cypher generation accuracy were calculated only for correctly classified queries to ensure that execution quality reflects successful task interpretation.

| Task Type | Query Set Size | Avg. Correct Task Classification | Avg. Valid Tool/Query Input | Extraction Accuracy (%) |
|---|---|---|---|---|
| Upward Reasoning | 20 | *19.8* | *19.2* | *97.0* |
| Downward Reasoning | 20 | *19.6* | *19.6* | *100.0* |
| Explanatory Query | 20 | *20.0* | *19.2* | *96.0* |

Table 2. LLM Agent Evaluation by Task Type Across 5 Runs

The LLM agent demonstrated consistent performance across all query types. Averaged over five independent runs of a 60-query test set, the agent achieved high classification accuracy, correctly identifying the intended reasoning or retrieval task in nearly all cases. For reasoning tasks, the agent achieved an argument extraction accuracy of 97.0% for upward reasoning, with an average of 19.2 correct extractions out of 19.8 correctly classified queries per run. For downward reasoning, both classification and extraction accuracy averaged 19.6 per run, corresponding to an extraction accuracy of 100.0%. For explanatory queries, the agent correctly classified all 20 queries per run on average and successfully generated Cypher queries for 19.2 of them, resulting in a 96.0% accuracy following correct classification. These results demonstrate the agent's reliability in distinguishing between diagnostic and interpretive tasks and its effectiveness in performing structured reasoning and knowledge retrieval based on the system model.

## 7. Conclusion and Outlook

### 7.1. Limitations and Challenges

Despite the demonstrated effectiveness of the proposed LLM- and KG-based diagnostic framework for automating DML model construction and enabling interactive fault analysis, several limitations remain. LLM outputs, while powerful, are varied and may generate hallucinated or incomplete structures even when guided by validation gates and the prompt-chaining workflow. This highlights the need for a human-in-the-loop process, where domain experts review and revise the automatically generated DML models to ensure logical consistency and domain accuracy. The framework also assumes access to well-structured documentation and reliable operational or historical data, which may not always be available in practice. Although the proposed architecture reduces manual effort, expert oversight remains essential. Furthermore, while the evaluation reports high element-level extraction accuracy, it does not capture the semantic impact of missing critical nodes. In DML models, elements such as gates, subfunctions, or success conditions are often essential to maintaining the integrity of fault propagation paths. The omission of even a single high-impact node can break logical chains and lead to incomplete or misleading diagnostics. This limitation suggests a need for broader evaluation strategies that assess whether generated models preserve full diagnostic reasoning capabilities. The current validation also relies on a curated query set based on typical expert interactions, which, while practical, may not reflect edge cases, ambiguous phrasing, or linguistic variation. More comprehensive testing involving adversarial queries, paraphrased inputs, and real-user feedback will be necessary to improve the robustness and generalizability of the system. Lastly, adapting the framework to domains beyond nuclear diagnostics may require customized prompts and tool modifications, which could limit its immediate applicability elsewhere.

Additionally, while the framework has been demonstrated on a moderately scoped system, its performance and scalability in large-scale, highly complex systems remain untested. The current case study involves a relatively simple subsystem with limited components and interactions. Future work should investigate how the framework performs when applied to large, mission-critical systems with deeply nested hierarchies, extensive dependencies, and real-time operational data. Understanding the computational requirements, performance bottlenecks, and accuracy trade-offs in such settings will be essential for broader adoption.

### 7.2. Conclusion

The integration of LLMs and KGs into diagnostic modeling marks a significant advancement in automating complex system analysis. This research introduces a scalable, AI-driven framework that streamlines the generation of structured diagnostic models and enhances predictive accuracy and fault reasoning through natural language interaction. By reducing reliance on manual modeling and enabling structured, explainable diagnostics, the approach adapts effectively to evolving system configurations. Depending on the query type, system knowledge from the KG is either processed through diagnostic tools or embedded into the LLM prompt, as previously described. This enables the LLM to generate responses that are both context-aware and grounded in system logic. The framework also facilitates human-AI collaboration by allowing users to interact with system behavior through intuitive queries, lowering the technical barrier to advanced diagnostics. It extends the utility of functional modeling techniques such as DML, which have traditionally required extensive domain expertise and manual effort. While demonstrated on a nuclear power application, the framework is broadly applicable to other mission-critical domains, including advanced manufacturing systems such as integrated circuit fabrication facilities. Designed for continual refinement using operational feedback and real-time data, the framework is positioned to support high-reliability industries with improved diagnostics, proactive risk management, and system resilience. The proposed framework was validated through comprehensive evaluations of both KG construction and LLM-based query interpretation. Across five independent runs, the system consistently produced accurate DML-based models, achieving over 90% extraction accuracy for critical logic structures such as gates and success conditions, and perfect identification of high-level goals. The LLM agent demonstrated strong performance across 60 diagnostic and explanatory queries, with high classification accuracy, reliable tool invocation, and consistent argument extraction. These results validate the framework's ability to generate interpretable, graph-based diagnostics directly from unstructured documentation. Overall, this work establishes a solid foundation for next-generation diagnostic systems that combine natural language interaction with structured reasoning. The implementation code and supplementary materials for the proposed framework are available at: *https://github.com/s-marandi/LLM-Based-Complex-System-Diagnostics*

### 7.3. Future Work

Future work will focus on advancing the evaluation and validation of automatically generated DML models. Building on the current framework, which includes multi-run assessments of extraction accuracy and hallucination rates, future evaluations will incorporate deeper semantic validation techniques. One direction involves benchmarking generated cut-sets and path-sets against expert-engineered baselines to assess both logical soundness and coverage. In addition to element-level accuracy, graph-level structural metrics such as connectivity fidelity, dependency correctness, and fault propagation traceability will be introduced. Evaluation will also extend to the robustness of LLM behavior under varied prompt conditions and alternative phrasings. Autonomous LLM agents will be further explored for iterative model refinement, leveraging feedback loops to enhance precision, reducing hallucinations, and adaptively correct errors over time by moving toward scalable, self-improving diagnostic model generation. Advanced diagnostic capabilities will also be developed, including probabilistic assessments of component criticality in cases where partial functionality can be tolerated. This will enable the framework to recommend optimal maintenance or mitigation strategies under uncertainty. As systems grow more complex, enhancements will focus on integrating real-time operational data and applying data fusion techniques to combine inputs such as sensor readings, maintenance records, and expert observations into the KG. To better model dynamic behavior, future efforts will introduce more sophisticated gating mechanisms capable of capturing time-dependent relationships and evolving system states. The model construction process will also be extended with advanced NER techniques to improve the precision and depth of information extracted from technical documentation. For systems with lengthy descriptions that exceed a single LLM context window, more effective text chunking and summarization strategies will be implemented, along with models that support larger token capacities. Together, these enhancements will further align the framework with the demands of real-time diagnostics in safety-critical, complex engineered systems.

# Appendix

**Step 1 Prompt**

```
prompt = f"""
Summarize the given system description for constructing a Dynamic Master Logic (DML)
by organizing the system's Goal, Functions, Subfunctions, Components, and Success Conditions
hierarchically using logical gates (AND/OR). Extract key relationships preserving the
dependency flow:
Goal → Gate -> Function → Gate -> Subfunction → Gate -> Component → Gate -> Success Condition.
Ensure all relationships are labeled (e.g., "ACHIEVED_BY", "DEPENDS_ON", "REQUIRES", "SUCCESS_THROUGH"). Provide missing information as 'N/A'.

System Description:
{system_description}

Return a structured response.
"""
```

**Gate 1 Prompt**

```
prompt = f""""

Review the following system summary. Check if it includes clearly defined goals, functions, subfunctions, components, and success conditions. Ensure that the logical flow from goal to components is present and coherent.

If any of these elements are missing, vague, or incomplete, respond with "Fail" and list the specific issues. If all elements are clearly present, respond with "Pass".

"""
```

Figure 9, Step 1 and Gate 1 Prompts

**Step 2 Prompt**

```
prompt = f"""
Convert the following structured system description into a strict JSON format.
Format:
```json
{{
  "Goal": {{
    "name": "Goal Name",
    "achieved_by": {{
      "gate": "AND_gate or OR_gate",
      "functions": [
        {{
          "name": "Function Name",
          "depends_on": {{
            "gate": "AND_gate or OR_gate",
            "subfunctions": [
              {{
                "name": "Subfunction Name",
                "requires": {{
                  "gate": "AND_gate or OR_gate",
}}... Similar formatting was followed but cropped due to spacing
```

**Gate 2 Prompt**

```
prompt = f"""

Validate the following JSON structure that represents a hierarchical system model. Check that:
- All required fields (goal, functions, subfunctions, components, success conditions) are included
- The nesting follows this order: Goal → Function → Subfunction → Component → Success Condition
- Logical gates (AND/OR) are correctly placed and labeled
- Keys are correctly named and formatted

Respond with "Pass" if the structure is valid. Respond with "Fail" and specify the structural or syntax issues if any.

"""
```

Figure 10, Step 2 and Gate 2 Prompts



| Step 3 Prompt | Gate 3 Prompt |
|---|---|
| prompt = f"""<br>    Convert the following JSON-based Knowledge Graph into **valid Cypher queries**<br>    using Neo4j syntax. Use the following **node types**:<br>    - `Goal`<br>    - `Function`<br>    - `Subfunction`<br>    - `Component`<br>    - 'Success Condition'<br>    - `AND_gate`<br>    - `OR_gate`<br><br>    For each node type pick the appropriate name for the node you are creating all the nodes types need a name.<br>    **Relationships:**<br>    - `ACHIEVED_BY` (Goal → Gate)<br>    - `DEPENDS_ON` (Gate → Function or Function → Gate)<br>    - `REQUIRES` (Gate → Subfunction or Subfunction → Component)<br>    - `SUCCESS_THROUGH` (Component gate → SuccessCondition) | prompt = f"""<br><br>Review the following Cypher queries intended to construct a knowledge graph. Check that:<br>- The syntax is correct for all CREATE and MATCH statements<br>- Node and relationship labels are properly quoted and capitalized<br>- The logical structure reflects the hierarchy from goal to success conditions<br>- No malformed or incomplete statements exist<br><br>Respond with "Pass" if the queries are syntactically and structurally valid. Respond with "Fail" and describe any problems found.<br><br>""" |

Figure 11, Step 3 and Gate 3 Prompts